\documentclass[times,final,longtitle]{elsarticle}

\usepackage{framed,multirow}
\usepackage{cite}
\usepackage{amssymb}
\usepackage{latexsym}
\usepackage{lscape}
\usepackage{url}
\usepackage{xcolor}

\usepackage{hyperref}
\usepackage{cite}
\usepackage{amsmath,amssymb,amsfonts}
\usepackage{algorithmic}
\usepackage{color}
\usepackage{graphicx}
\usepackage[misc,geometry]{ifsym} 
\usepackage{multirow}
\usepackage[ruled,linesnumbered]{algorithm2e}
\graphicspath{ {./figures/} }
\usepackage{booktabs}
\usepackage{tablefootnote}
\usepackage{tikz}
\usetikzlibrary{bayesnet}
\usepackage{pgfplots}
\usepackage{wrapfig}
\usepackage{graphicx}
\usepackage{textcomp}
\usepackage{soul}
\usepackage{todonotes}
\usepackage{color}
\usepackage{multirow}

\SetCommentSty{mycommfont}

\newcommand{\shape}{\mathcal{S}}
\newcommand{\shapet}{\tilde{\mathcal{S}}}
\newcommand{\pos}{\mathbf{x}}

\newcommand{\lref}{l_{\mathrm{ref}}}

\newcommand{\lopt}{l_{\mathrm{opt}}}
\newcommand{\sdm}{\mathrm{SDM}}

\newcommand{\params}{\boldsymbol{\theta}_S}
\newcommand{\aka}{{a.k.a.}~}

\newcommand{\tpdf}{t}
\newcommand{\fg}{f_g}
\newcommand{\bg}{b_g}

\newcommand{\euc}{\mathbb{R}}
\newcommand{\sigmoid}{\sigma}
\newcommand{\rot}{\mathbf{R}}
\newcommand{\rotv}{\mathbf{r}}
\newcommand{\trans}{\mathbf{t}}

\newcommand{\identity}{\mathbf{I}}

\newcommand{\paramsr}{\boldsymbol{\theta}_{SR}}
\newcommand{\paramsd}{\boldsymbol{\theta}_{SD}}
\newcommand{\hatparams}{\hat{\boldsymbol{\theta}}_S}
\newcommand{\parami}{\boldsymbol{\theta}_I}

\newcommand{\hatparami}{\hat{\boldsymbol{\theta}}_I}

\newcommand{\paramd}{\theta_D}

\newcommand{\priors}{\boldsymbol{\alpha}}
\newcommand{\priori}{\boldsymbol{\beta}}

\newcommand{\meanpv}{{\boldsymbol{\mu}}}

\newcommand{\datad}{\mathbf{d}}
\newcommand{\cov}{\boldsymbol{\Sigma}}
\newcommand{\grad}{\mathbf{g}}
\newcommand{\hess}{\mathbf{H}}
\newcommand{\hesst}{\tilde{\boldsymbol{H}}}

\DeclareMathOperator{\diag}{Diag}

\definecolor{newcolor}{rgb}{.8,.349,.1}

\journal{Medical Image Analysis}

\usepackage{natbib}

\begin{document}
\let\WriteBookmarks\relax
\def\floatpagepagefraction{1}
\def\textpagefraction{.001}

\title{Bayesian Logistic Shape Model Inference : application to  cochlea image segmentation
}%

\author[1,2]{Wang Zihao}
\cortext[1]{Corresponding author:}\ead{zihao.wang@inria.fr}
\author[4]{Demarcy {Thomas}}
\author[2,3]{Vandersteen Clair}
\author[4]{Gnansia {Dan}}
\author[2,5]{Raffaelli Charles}
\author[2,3]{Guevara Nicolas}
\author[1,2]{Delingette Herv\'{e}} 

\address[1]{Inria Sophia Antipolis Méditerranée,  2004 Route des Lucioles,  06902 Valbonne, FRANCE}
\address[2]{Université Côte d'Azur, 28 Avenue de Valrose, 06108 Nice, FRANCE}
\address[3]{Head and Neck University Institute, Nice University Hospital, 31 Avenue de Valombrose, 06100 Nice, FRANCE}
\address[5]{Department of Radiology, Nice University Hospital, 31 Avenue de Valombrose, 06100 Nice, FRANCE}
\address[4]{Oticon Medical, 14 Chemin de Saint-Bernard Porte, 06220 Vallauris, FRANCE}

\begin{abstract}
Incorporating shape information is essential for the delineation of many organs and anatomical structures in medical images. While previous work has mainly focused on parametric spatial transformations applied on  reference template shapes,  in this paper, we address the Bayesian inference of parametric shape models for segmenting medical images with the objective to provide  interpretable results. The proposed framework defines a likelihood appearance probability and a prior label probability based on a generic shape function through a logistic function.  A reference length parameter defined in the sigmoid  controls the trade-off between shape and appearance information.
The inference of shape parameters is performed within an Expectation-Maximisation approach where a Gauss-Newton optimization stage allows to provide
an approximation of the posterior probability of shape parameters. 

This framework is applied to the segmentation of cochlea structures from clinical CT images constrained by a 10 parameter shape model. It is evaluated on three different datasets, one of which includes more than 200 patient images. The results show performances comparable to supervised methods and better than previously proposed unsupervised ones. It also enables an analysis of  parameter distributions and the quantification of segmentation uncertainty including the effect of the shape model.  
\end{abstract}

\begin{keyword}
Bayesian Inference \sep Image Segmentation  \sep Shape Modeling
\end{keyword}
\maketitle


\section{Introduction}
\label{sec:introduction}
Several anatomical structures have a typical shape, such that a medical expert can easily recognize them from their three-dimensional representation. This is for instance the case of basal ganglia within the brain~\citep{ASHBURNER2005839}, but also of abdominal structures, such as the liver or  kidneys. Another emblematic example is the cochlea which is a small organ within the inner ear having a remarkable spiraling configuration where  mechanical waves are transformed into electrical stimulation of the auditory nerve. The cochlea shape is complex as it completes around two and a half turns  with its centerline closely resembling a logarithmic spiral helix \citep{Cohen1996, Baker2008}. Its segmentation from CT images of the temporal bone is challenging since those  images  have low resolution with respect to the anatomy of the cochlea: the cochlea dimension is about 8.5x7x4.5~mm$^3$ while the typical CT voxel size is larger than 0.2~mm which is  weakly visible for the fine structures of the chambers.In addition, the cochlea is filled with fluids that can be  found in the vestibular system and other neighbouring structures, with similar appearance in CT images.  


Supervised learning (e.g. Deep Learning)  is  an  effective way to perform image segmentation or processing in many cases. Specifically, in inner ear CT imaging analysis, many works achieved impressive results \citep{DICE90CNN,DICEUNET90,DICE90,MARGAN,li2021microct,s19194139,Zhang,One_shot_co}. However, supervised learning methods have also many limitations. First, creating  dataset annotations is time consuming, possibly preventing the creation of massive training datasets. In the cochlea case, a well trained ENT surgeon would  need at least ten minutes to segment each 3D cochlea volume. Second, due to the potential  overfitting related to the limited training set, the output of such supervised algorithm  is  likely to fall outside the shape space of the structure of interest.

Shape-based image segmentation can overcome the above limitations since the optimization of the model can be done in an unsupervised or weakly supervised way. Besides, the recovered shape parameters make  a natural compact representation that is useful for shape analysis and even clinical applications.
 In this paper, we consider  shapes that are either defined as an explicit $\shape(\params)\in\euc^d$ or  implicit $\shape(\params,\pos)=0$ parametric shape models where $\params$ is a set of shape parameters and $\pos\in\euc^d$,  is any point in space ($d=2,3$).

Those parametric shape models  serve to guide the delineation of such anatomical structures by constraining the shape space of the segmented object. We can 
roughly split the shape-based image segmentation methods into two sets of methods. A first set 
optimizes the shape parameters $\params$  by minimizing the sum of a regularizing term $E_R(\params)$ and an image term $E_I(\shape(\params),I,\parami)$ : 
$\hatparams=\arg \min_{\params} E_I(\shape(\params),I,\parami)+E_R(\params)$
where $\parami$ is a set of image parameters that may also be optimized. 
This iconic shape fitting principle is typically used in the classical active shape model~ \citep{Cootes1995,Heimann2007} and their extensions~\citep{Cremers-et-al-03hd}.
Various generic image terms may be considered for instance as those explored in~\citep{Tsai2003}. 
A second set of methods uses the  shape model $\shape(\params)$ as a shape prior instead of a shape space. Several shape constraints have been introduced within several image segmentation frameworks including  level-sets~\citep{1467575,Cremers03}, free-form deformation space~\citep{1216225} or implicit template deformation~\citep{Prevost2013}.
While those methods have greater shape flexibility  for delineating structures, it is often difficult to set the coefficients weighting the shape constraint with other image terms. 
Those two sets of shape based segmentation methods are expressed as energy minimization problems, thus  only allowing  to have point estimates of shape parameters and not their posterior probabilities. 

Another common shape representation consists in specifying a parametric spatial transformation $\mathcal{T}(\paramd):\euc^d\rightarrow \euc^d$ acting on a template shape $\shape(\theta_0)\in\euc^d$ leading to an indirect shape parameterization : $\shape(\paramd)=\mathcal{T}(\paramd)\circ\shape(\theta_0)$. This  formulation of shape modeling based on a deformable template  leads to solving  a joint segmentation and registration problem. More precisely, several authors~\citep{Ashburner2005,Pohl2006} defined generative image and shape models and performed statistical variational inference to optimize their parameters and  hyperparameters. 
Priors on the deformation space based for instance on minimal  elastic energy~\citep{4695995}, were applied on  triangular or tetrahedral mesh templates. 
Other shape priors were defined as restricted Boltzmann machines~\citep{AGN2019220} or as shape-odds~\citep{8099718}. In most cases, optimal shape parameters (e.g. mesh vertex positions) are obtained as maximum a posteriori but not their posterior probability. Uncertainty quantification of image registration algorithms has been tackled in some research papers~\citep{SIMPSON20122438, wang_uncer, LOIC} based on a low dimensional representation of deformation space and Laplace approximation.

In this paper, we propose a novel Bayesian framework for shape constrained image segmentation based on parametric shape models (instead of parametric spatial transformations) where   the output segmentation is driven by a shape model but without restricting it to a low dimensional space. 
The proposed approach is generic as it is suitable for 
any explicit $\shape(\params)$ and implicit $\shape(\params,\pos)=0$ parametric shape models associated with any appearance models representing the intensity distributions inside background or foreground regions.
It is based on a logistic shape  prior  defined as the sigmoid of a  shape function (e.g. signed distance map) defined over the image domain. Inferences of shape and intensity parameters are performed by maximizing the joint image and shape parameters probability $p(\params,\parami,I)$ with an Expectation-Maximization algorithm. We show that this optimization boils down to 
having the posterior label distribution as close as possible (in terms of Kullback Leibler divergence) from both the likelihood and shape prior distributions.
A Gauss-Newton optimization method is introduced to optimize the shape parameters leading to closed form updates similarly to  iterative reweighted least squares schemes.  It outputs the most probable shape and imaging parameters but also an approximation of the posterior shape parameter probability which is essential for estimating the segmentation uncertainty.

This framework is applied to the  problem of cochlea segmentation on CT images based on a parametric shape model with 10  parameters, and an imaging model defined as a mixture of Student's $t$-distributions. It results in the reconstruction of cochlea structures in 2 small datasets consisting of paired  CT and $\mu CT$ post-mortem images and one large dataset of nearly 200  patient CT images.  We showed that the proposed framework leads to state of the art reconstruction performances as well as the recovery of consistent shape parameter distributions and the estimation of segmentation uncertainty.  

The main contributions of this article are:
\begin{itemize}
    \item[-] A novel framework for image segmentation that combines probabilistic appearance and shape models. It is generically defined for  parametric shape functions rather than parametric space transformations.  The trade-off between the appearance and shape models is governed by an interpretable parameter : the reference length.
    \item[-] A Gauss-Newton optimization method of the shape parameters which also produces a posterior approximation of those shape parameters. 
    \item[-] A method for uncertainty quantification of image segmentation which takes into account the shape uncertainty.
    \item[-] A segmentation method of the cochlea in clinical CT images which  provides state-of-the-art results and interpretable shape parameters.
\end{itemize}


We present below the framework of the logistic shape model (section~\ref{sec;method}), the shape and intensity  models used specifically for cochlea segmentation (section~\ref{sec;application-cochlea}), and the segmentation results on 3 clinical and pre-clinical datasets (section~\ref{sec;results}).

\section{Method}
\label{sec;method}
\subsection{Shape-based Generative Probabilistic Model}
We consider an observed image $I$ consisting of N voxels  $I_n\in\euc$, $n = 1, \dots, N$, for which we seek to solve a binary segmentation problem guided by a  shape model. That model is defined either as in a parametric form as $\shape(\params)\in\euc^d$, $d=2,3$ or in an implicit form as $\shape(\params,\pos)=0$. In the case of parametric shape models, one can define an associated implicit function $\sdm(\shape(\params),\pos)=0$ as the signed distance map defined at point $\pos$. Therefore, we propose to unify notations for both parametric and implicit cases by stating  the existence of   a {\em shape function} $\shapet(\params,\pos)\in\euc$ whose zero level defines a shape and whose sign indicates if a point is inside (positive) or outside (negative). Note that with this hypothesis,  a shape corresponds to a (smooth) manifold of co-dimension 1 without borders, thus defining a partition of the image into inside and outside regions.  


A binary label variable $Z_n\in\{0,1\}$ is defined at each voxel  specifying if voxel $n$ belongs to the background or foreground regions. A probabilistic intensity  distribution model  is defined for each region $p(I_n|Z_n=k,\parami^k)$, $k=0,1$ controlled by the intensity parameter array  $\parami^k$. 
The arrays for background ($k=0$) and foreground ($k=1$) are concatenated into the intensity parameter array $\parami$. This appearance model can be either supervised , e.g. a trained convolutional neural network, or unsupervised, e.g. a Gaussian mixture model. In the remainder, we assume the latter case and therefore we define mechanisms to optimize  the appearance parameters $\parami$. In the supervised case, the steps involving the update of $\parami$ should be ignored.  

We enforce a spatial correlation between the label of each voxel  by specifying  their \emph{a priori} dependence on the shape model $\shapet(\params,\pos)$.   More precisely, we define a prior probability for voxel $n$ to belong to the foreground region as follows:
\begin{align}
p(Z_n=1|\params)&=\sigmoid\left(\frac{\shapet(\params,\pos_n)}{\lref}\right)\label{eq;prior} \\ p(Z_n=0|\params)&= 1-p(Z_n=1|\params)=\sigmoid\left(-\frac{\shapet(\params,\pos_n)}{\lref}\right)
\nonumber
\end{align}
where $\sigmoid(x)$ is the sigmoid (or logistic) function, $\pos_n$ is the position of voxel $n$ and $\lref$ is a reference length. With that definition, the prior probability will be close to 1 inside the object, close to 0 outside and equal to 0.5 on the shape boundary. We call this formulation of the label prior,  the \emph{logistic shape model} as it combines shape information into a probability distribution through a logistic function. This definition of the shape prior is related to several prior work in the literature such as 
  probabilistic atlases and LogOdds maps~\citep{Pohl2006a}, continuous STAPLE~\citep{continuousstaple}, a nd label fusion~\citep{labelfusion}.

The quantity $\lref$ is a characteristic length  which controls the slope of the  prior probability next to the object boundary. This parameter also influences the trade-off between intensity and shape information in the segmentation process as discussed in section~\ref{subsec;lref}.
The  shape  parameters $\params$ are themselves regarded as random variables with a multivariate Gaussian prior controlled by hyper-parameters $\priors$: $p(\params|\priors)$. The intensity parameters may also optionally be considered as random variables with hyper-parameter $\priori$ as  $p(\parami|\priori)$. The shape based generative model is summarized in Fig.~\ref{fig:segmentationRes}:(a).

\subsection{Logistic Shape Model Framework }

With the proposed generative model, given an image, the objective is to infer the most probable values of the intensity $\hatparami$  and shape parameters $\hatparams$ which will lead to the  estimation of  the posterior label probabilities given by :
\begin{align}
p(Z_n=1|I_n,\parami,\params)&=\frac{p(I_n|Z_n=1,\parami^1)p(Z_n=1|\params)}{\sum_{k=0}^1 p(I_n|Z_n=k,\parami^1)p(Z_n=k|\params)}
\end{align}
That posterior probability is clearly a compromise between shape information stored in the prior $p(Z_n=1|\params)$ and appearance information stored into the likelihood $p(I_n|Z_n=1,\parami^1)$. The \emph{segmented region of interest} (SROI) then corresponds to voxels for which $p(Z_n=1|I_n,\parami,\params)\geq \frac{1}{2}$. In addition, the logistic shape model framework recovers the most likely shape parameters $\hatparams$ that corresponds to the \emph{segmented shape instance} (SSI) which is the best fit of the shape model in that image. Finally, we will show that we can approximate the posterior shape parameter $p(\params|I)$ in order to capture the uncertainty in the shape parameter estimation.

The optimization of the intensity and shape parameters is done by maximizing the the log-joint intensity and parameters  probability :
\begin{equation}
\begin{aligned}
(\hatparams,\hatparami) &= \arg \max_{\params,\parami}  \log p(I,\params,\parami)= \arg \max_{\params,\parami} \mathcal{L}(\params,\parami)\\
\mathcal{L}(\params,\parami)&=  \log p(I|\params,\parami)+\log p(\params)+\log p(\parami)\\
&= \sum_{n=1}^N \log \left(\sum_{k=0}^1 p(I_n|Z_n=k,\parami) ~p(Z_n=k|\params)\right)+\\
&\log p(\params)+\log p(\parami)  
\label{eq:logp} 
\end{aligned}
\end{equation}

In the log-joint probability $\mathcal{L}(\params,\parami)$ we have marginalized out the hidden label variables $Z_n$ and used  the conditional independence of variables $I_n$ given $\params$. 

\subsection{Expectation-Maximization Inference}
The direct optimization of $\mathcal{L}(\params,\parami)$ can be done by any optimization toolbox but it is difficult due to the possible encountered overflows/underflows caused by the log-sum-exp expressions. 



This is why we propose to follow the  Expectation-Maximization (EM) algorithm
which  relaxes that optimization problem into several optimizations  over simpler problems. We proceed by introducing 
 N variables $u_n$ that are surrogates for  the posterior label probability $p(Z_n=1|I_n,\params,\parami)$ such that  $u_n\in[0,1]$. Writing $U=\{u_n\}$, we introduce a new augmented criterion $\mathcal{L}^*(\params,\parami,U)=\log p(I,\params,\parami)-D_{\mathrm{KL}}(U||p(Z|I,\params,\parami))$ by adding the negative Kullback-Leibler divergence between $u_n$ and the posterior label $p(Z_n|I_n,\params,\parami)$.
 
Maximizing $(\params,\parami,U)$ over the augmented criterion $\mathcal{L}^*(\params,\parami,U)$ leads to the same optima in $(\params,\parami)$ than the maximization of  $\mathcal{L}(\params,\parami)$ but with simpler expressions:
\begin{equation*}
\begin{aligned}
\mathcal{L}^*(\params,\parami,U)
&=\sum_{n=1}^N \sum_{k=0}^1 u_n^k \log ( p(I_n|\params,\parami) p(Z^k_n=k|I_n,\params,\parami))\\
&-\sum_{n=1}^N \sum_{k=0}^1  u_n^k\log u_n^k  +\log p(\params)+\log p(\parami) \\
&= Q(U,\params,\parami)+\sum_{n=1}^N H(u_n)+\log p(\params)+\log p(\parami)
\end{aligned}
\end{equation*}
where $Q(U,\params,\parami)=\mathbb{E}_U(\log p(I,Z|\params,\parami))$ is the conditional expectation of the complete marginal log-likelihood (\aka evidence) and $H(u_n)$ is the entropy of variable $u_n$. The quantity $Q(U,\params,\parami)$ is a lower bound of the log-likelihood since $H(u_n)>0$.

The maximization of the augmented criterion $\mathcal{L}^*(\params,\parami,U)$ is performed by the successive maximization over the $U$, $\parami$ and $\params$ variables. 
The {\bf E-step} corresponds to the maximization of $\mathcal{L}^*(\params,\parami,U)$ with respect to $U$ which  sets the surrogate variable $U$ to the posterior label probability $u_n=p(Z_n=1|I_n,\params,\parami)$.

The {\bf MI-step } optimizes  the log-joint probability with respect to  the appearance variables $\parami$, which is  equivalent to the maximization  of $\mathcal{L}_I=-D_{\mathrm{KL}}(U||p(I|Z,\parami))+\log p(\parami|\priori)$.  When the appearance parameters are independent between classes,  then $\log p(\parami|\priori)=\sum_{k=0}^K \log p(\parami^k|\priori_k)$ and the MI-step splits into $2$ independent maximization over $\parami^k$, $k=0,1$ of  $\mathcal{L}_I^k= -\sum_{n=1}^N D_{\mathrm{KL}}(u_n^k||p(I_n|Z_n=e_k,\parami^k))+\log p(\parami^k|\priori_k)$. For certain well chosen intensity models such as Gaussian mixture models, this optimization leads to closed-form updates of $\parami$.

Finally, we perform the {\bf MS-step } corresponding to the maximization over shape variables $\params$  which  is equivalent to the maximization of $\mathcal{L}_S$:
 \begin{equation*}
 \begin{aligned}
 \mathcal{L}_S&= -D_{\mathrm{KL}}(U||p(Z|\params))+\log p(\params|\priors) \label{eq;MS} 
  \end{aligned}
 \end{equation*}
 
We can see that the EM algorithm preserves  an interesting symmetry between shape and appearance information. Indeed, the iterative application of the E, MS and MI steps makes the posterior labels distribution $U$ as close (in terms of KL divergence) as possible from the likelihood $p(I|Z,\parami)$ and shape prior $p(Z|\params)$ that the minimization of $D_{\mathrm{KL}}(U||p(Z|\params))+D_{\mathrm{KL}}(U||p(I|Z,\parami))$. At convergence, the posterior distribution is therefore clearly a compromise between shape and appearance information.



\subsection{Optimization of shape parameters \texorpdfstring{$p(\params|I)$}{} }

The functional $\mathcal{L}_S$ is a non trivial function of the parameters $\params$ as it combines 2 non-linear functions : 
the sigmoid $\sigmoid()$ and the shape function $\shapet(\params,\pos_n)$: 
\begin{equation}
    \begin{split}
    \mathcal{L}_S&= -\sum_{n=1}^N \left (u_n \log \sigmoid\left(\frac{\shapet(\params,\pos_n)}{\lref}\right)\right. +
    \left. (1-u_n)\log \sigmoid\left(-\frac{\shapet(\params,\pos_n)}{\lref}\right)\right) \\
  & + \log p(\params|\priors)  +\mathrm{cst}  \label{eq:LS}
    \end{split}
\end{equation}
The  functional gradient $\nabla_{\params}\mathcal{L}_S$ cannot  be written in closed form since  it    requires  the computation of the 
 gradient of the scaled shape function at each voxel~: $\datad_n=\frac{\nabla_{\params}\shapet(\params,\pos_n)}{\lref}\in\euc^{|\params|}$. Those gradient vectors may be computationally costly to compute, for instance when the shape function is based on a signed distance map of parametric shape models $\shapet(\params,\pos)=\sdm(\shape(\params),\pos)$. In that case, the $\datad_n$ values are computed by a costly finite difference approximation except for translation and rotation parameters for which they can be
  computed efficiently (see \ref{appendixgradientshape}). 
   After combining all $\datad_n$ terms in a gradient matrix $\datad\in\euc^{|\params|\times N}$, the functional gradient can be simplified as $\nabla_{\params}\mathcal{L}_S= -\datad(\mathbf{u}-\meanpv)+\nabla_{\params} \log p(\params|\alpha)$ where $\mathbf{u}=(u_1 \ldots u_N)^T\in\euc^N$ and  $\meanpv=\left (\sigmoid\left(\frac{\shapet(\params^i,\pos_1)}{\lref}\right)\ldots \sigmoid\left(\frac{\shapet(\params^i,\pos_n)}{\lref}\right)\right)^T\in\euc^N$.
 
Thus, a first approach for optimizing the shape parameters is to use any  quasi-Newton optimization method such as the BFGS algorithm ( similarly to~\citep{demarcy:tel-01609910}),  since it only requires the computation of the functional gradient and iteratively estimates the Hessian matrix. Yet, this generic optimization was found to be fairly time consuming and sometimes unstable. 

Instead,  we propose to adopt a Gauss-Newton optimization approach where we approximate the Hessian matrix by ignoring the term involving second order derivatives. More precisely, the Hessian of the functional is computed as $\hess= \nabla^2_{\params}\mathcal{L}_S= -\nabla_{\params}\datad\otimes (\mathbf{u}-\meanpv)-\datad\otimes \nabla_{\params}\meanpv + \nabla^2_{\params}\log p(\params|\alpha)$. After dropping the first term, we get the following approximate Hessian $\hess\approx\hesst= -\datad\otimes \nabla_{\params}\meanpv + \nabla^2_{\params}\log p(\params|\alpha)$. When inserting the expression of the gradient of the prior, we get : $\hesst=\datad \diag(\meanpv\circ(1-\meanpv))~\datad^T+ \nabla^2_{\params}\log p(\params|\alpha)$ where $\circ$ is the element-wise product between two vectors. This approximate Hessian matrix is positive definite by construction and is then used to perform several Newtons steps. 

The sketch of the MS step is shown as algorithm~\ref{alg:msstep} where  the shape parameter prior $p(\delta\params^i|\priors)$ is arbitrarily chosen as a zero mean Gaussian distribution with covariance $\Sigma^0_{\params}$. It consists of  two intertwined loops, the innermost performing iteratively the Newton updates and updating the mean, gradient and Hessian values. The outer loop updates the shape function gradient which is potentially a costly step. In line 15 of the algorithm,  the $U$ variable is updated in an E-step in order to speed-up the convergence of the overall EM algorithm. Since the parameter range is bounded, we perform in practice a truncated Newton step as proposed in~\citep{Method_bounded}.

This Gauss-Newton approach was inspired by the iterative re-weighted least squares algorithm~\citep{bishopbook} developed for solving logistic regression (LR) problems. Indeed the first term of $\mathcal{L}_S$ is similar to the log likelihood of LR  after  replacing $u_n$ with a binary variable and linearizing the shape function. 
The proposed approach is also related to the Fisher scoring algorithm (see \citep{fishermedical} as an example in medical image analysis) when the point-wise Hessian matrix of the log likelihood  is replaced by its expectation thus leading to more stable evaluation. In this particular case, the approximate Hessian is not the expectation of the Hessian since the first term of  $\mathcal{L}_S$ is the expectation of the log-prior with respect to binary variable $U$ instead of $Z$.

 Finally, the proposed algorithm also outputs a Laplace approximation of the shape parameter posterior $p(\params|I)$ as a Gaussian distribution where the mean is the optimized shape parameter $\params^{\star}$ and the covariance is the inverse approximate Hessian matrix $\cov_{\params}^{\star}=(\hesst)^{-1}$. 
 
 The overall optimization finally  consists in iterating a series of outer loop, each loop consisting in optimizing the shape parameters as in Alg.~\ref{alg:msstep} then followed by a series of MI-steps until the relative change of intensity parameters is less than a threshold. The stopping criterion for the outer loop is 
 the relative change of foreground intensity parameters as it is the most impactful parameter. 

\begin{algorithm}[!h]
\SetKwFunction{MS Step}{MS Step}
 	$i\leftarrow 0$\;
    $u_n=p(Z_n=1|I,\params)$\tcp*[r]{E-Step, Update U}
 	\Repeat{ $\|\delta\params^{t+1}\|/\|\params^{t+1}\|<\epsilon$ 	}{
 	 	$\boldsymbol{V}\leftarrow\frac{\shapet(\params^i,\pos_n)}{\lref}\in\euc^{N}$\tcp*[r]{Shape function}
 	$\datad\leftarrow\frac{ \nabla_{\params}\shapet(\params^i,\pos_n)}{\lref}\in\euc^{|\params|\times N}$\tcp*[r]{Shape function gradient}
 	$\delta\params^0\leftarrow\mathbf{0}$, $t\leftarrow 0$\;
    \Repeat{ $\|\delta\boldsymbol{\theta}\|/\|\boldsymbol{\theta}\|<\epsilon$}{
    	$\meanpv\leftarrow\sigmoid(\boldsymbol{V+\datad^T\delta\params^t})$\tcp*[r]{Current Prior probability}
    $\grad\leftarrow -\datad(\mathbf{u}-\meanpv) -(\cov^0_{\params})^{-1}\delta\params^t$\tcp*[r]{Functional Gradient}
     $\hesst\leftarrow\datad \diag(\meanpv\circ(1-\meanpv))~\datad^T-(\cov^0_{\params})^{-1}$\tcp*[r]{Approximate  Functional Hessian}
     $\boldsymbol{\Sigma }^{\star}\leftarrow(\hesst)^{-1}$\tcp*{Covariance}
     $\delta\boldsymbol{\theta}\leftarrow -\boldsymbol{\Sigma }^{\star}\grad$\tcp*[r]{Truncated Gauss Newton Update}     $\delta\params^{t+1}\leftarrow\delta\params^{t}+\delta\boldsymbol{\theta}$, 
      $t\leftarrow t+1$\tcp*{Update  shape parameters}
    }
    $u_n=p(Z_n=1|I,\params)$\tcp*[r]{E-Step, Update U}
    $\params^{i+1}\leftarrow\params^{i}+\delta\params^{t+1}$,
     $i\leftarrow i+1$\tcp*[r]{end inner loop}
 	} 
    $\params^{\star}\leftarrow\params^{i}$, 
 	$\cov_{\params}^{\star}=\boldsymbol{\Sigma }^{\star}$\tcp*{Gaussian posterior}
 	\caption{MS step to compute $p(\params|I)$ }
 	\label{alg:msstep}
\end{algorithm}




\subsection{Influence of the characteristic length \texorpdfstring{$\lref^k$}{lrefk}}
\label{subsec;lref}


Based on Eq.\ref{eq;postlabel} and Eq.\ref{eq;prior}, it is easy to see that for infinitely small value of the characteristic length $\lref\rightarrow 0$, then the label prior becomes more and more sharp  $p(Z_n=1|\params)\rightarrow \delta_{\sdm(\shape(\params),\pos)>0}$ and the label posterior becomes equal to the label posterior : $p(Z_n = 1|\params,\parami,I_n)\longrightarrow p(Z_n=1|\params)$. Conversely, for infinitely large  value of the characteristic length  $\lref\longrightarrow \infty$, the label prior becomes uninformative $p(Z_n=1|\params)\longrightarrow \frac{1}{2}$ and the label posterior converges towards the appearance driven label posterior :  $p(Z_n = 1|\params,\parami,I_n)\longrightarrow p(I_n|Z_n = 1,\parami^1)/(p(I_n|Z_n = 0,\parami^0)+p(I_n|Z_n = 1,\parami^1))$.  Therefore the characteristic length controls the relative influence of the shape and appearance information in the probability of assigning a label. 

\begin{figure}
\centering
\begin{tikzpicture}
      \begin{axis}[height=0.34\textwidth,ylabel={$\mathbb{E}(p(Z_n = e_1|\theta_S,\theta_I,I_n))$}, xlabel={$\frac{d_n}{l_{\mathrm{Ref}}}$}, axis lines= middle,domain=-15:15,ymin=0,ymax=1, samples=50]
 \addplot[blue, ultra thick] {(1-x*exp(-x)-exp(-x))/(exp(-x)-1)^2};
\end{axis}
    \end{tikzpicture}
\caption{Expected label posterior probability as function of the normalized signed distance from the reference shape.}
\label{fig:postlabel}
\end{figure}
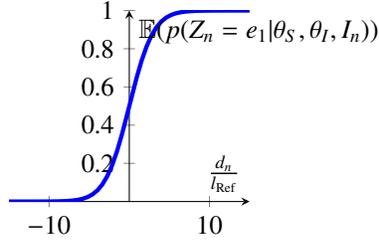

Since it is scaling the signed distance function, $\lref$ can be interpreted as controlling how far the  resulting shape given by $p(Z_n = e_1|\params,\parami,I_n)=0.5$ 
is allowed to deviate from the reference shape given by $\shape(\params)$. More precisely, assuming a uniform distribution of the appearance label probability between 0 and 1, one can compute the expectation of the posterior probability for a voxel located as a distance $d_n$ from the reference shape~: 
\begin{equation*}
\begin{aligned}
\mathbb{E}(p(Z_n = 1|\params,\parami,I_n))&=\int_0^1 \frac{t S(\Delta_n) }{t S(\Delta_n) +(1-t)(1- S(\Delta_n))}~dt\\
&= \frac{1-\Delta_ne^{-\Delta_n}-e^{-\Delta_n}}{(e^{-\Delta_n}-1)^2}~~~~,~~ \Delta_n=\frac{d_n}{\lref}
\end{aligned} 
\label{eq;postlabel}
\end{equation*}
Based on the graph of Fig.\ref{fig:postlabel}, a voxel located at least at $4\lref$ inside the boundary of the reference shape $\shape(\params)$ ($p<-4$) will have in average at least 95\% probability to be classified as belonging in the object. 

\section{Application to Cochlea Shape Recovery}
\label{sec;application-cochlea}

\subsection{Cochlea shape model}
\begin{figure*}[ht!]
    \centering
    \includegraphics[width=0.9\textwidth]{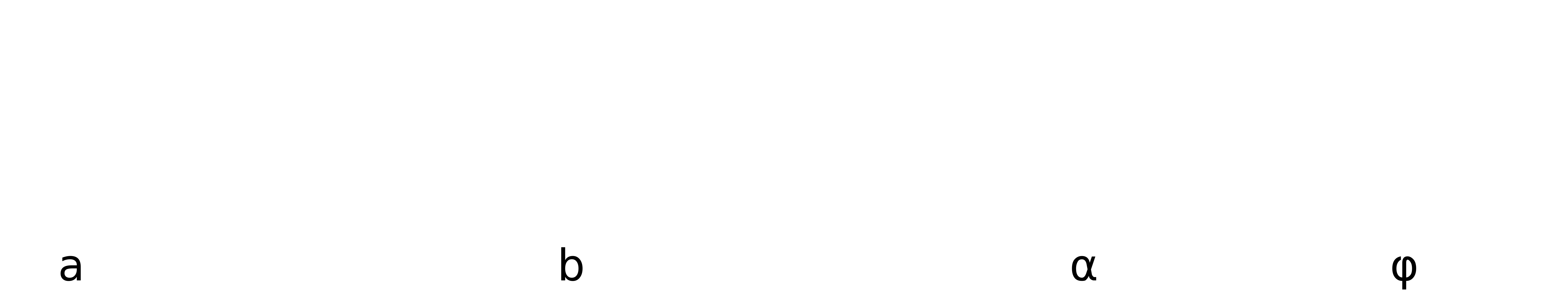}
    \caption{Parametric shape model of the cochlea. (Left)   Effect of the radial parameters $a$ (red), and $b$ (yellow) are shown with the reference position in purple; (Right) Effect of the longitudinal parameters $\alpha$ (pink) and  $\varphi$ (blue) parameters.}
    \label{fig:deformation}
\end{figure*}

We use a parametric cochlea shape model which is controlled by a set of 4 deformable shape parameters $\paramsd : \{a, b, \alpha, \varphi\}$ as shown in Fig:(\ref{fig:deformation}). Those 4 parameters control the deformation of the centerline of the cochlea represented as a generalized cylinder and is detailed in~\ref{appendixShape}. In addition to those 4 {\em deformable } parameters, we consider the 6 pose parameters $\paramsr$ consisting of rotation  $\{rx, ry, rz\}$ (parameterizing a rotation vector) and translation  $\{tx, ty, tz\}$ values. Therefore, the total number of shape parameters is 10, controlling the rigid and non rigid (deformable) motion: $\params=\paramsd\cup\paramsr$. 

To fit in our framework,  signed distance map $\sdm(\shape(\params),\pos)$ from the cochlea triangular mesh surface must be created. This can be performed for instance by using VTK functions~\citep{Maurer03alinear} but   
that distance map generation may take several seconds on large volumetric images. This is why  we have developed  a convolution neural network, noted as DLSDM, which outputs  an approximation of the signed distance map from the set of deformable shape parameters in few milliseconds on CPU~\citep{wang2020}. 
\begin{figure*}[h!]
\centering
    \includegraphics[width=\textwidth]{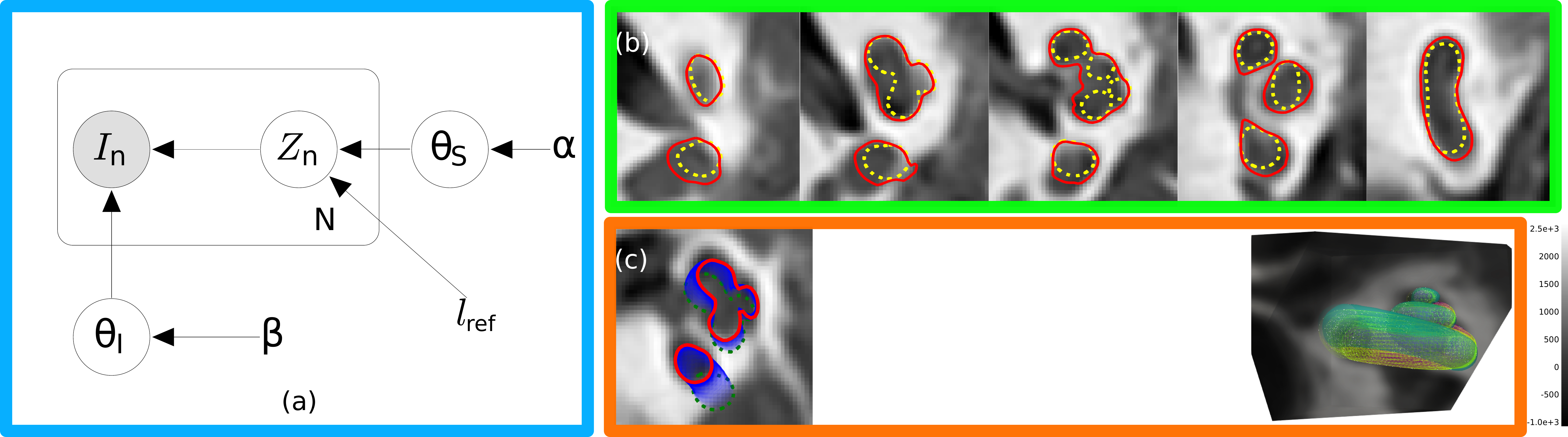}
    \caption{(a) graphical model for the shape-based generative model; (b) Cochlea segmentation on CT images is shown in solid red with the associated shape model in dashed yellow lines; (c) Evolution of the cochlea shape model during several MS steps shown as  2D contours (from dotted green to solid red) and 3D models. 
    }
    \label{fig:segmentationRes}
\end{figure*}

\subsection{Cochlea Appearance model}
Appearance models describe the intensity patterns inside the foreground and the background classes and can be built in a supervised, semi-supervised or unsupervised manner. Many simple  generative  models such as Gaussian mixture models (GMM)  with spatial corrections~\citep{Pohl2006, Ashburner2005} have been proposed in the literature to describe tissue intensity distributions.
\begin{figure}
	\centering
	\includegraphics[width=1\linewidth]{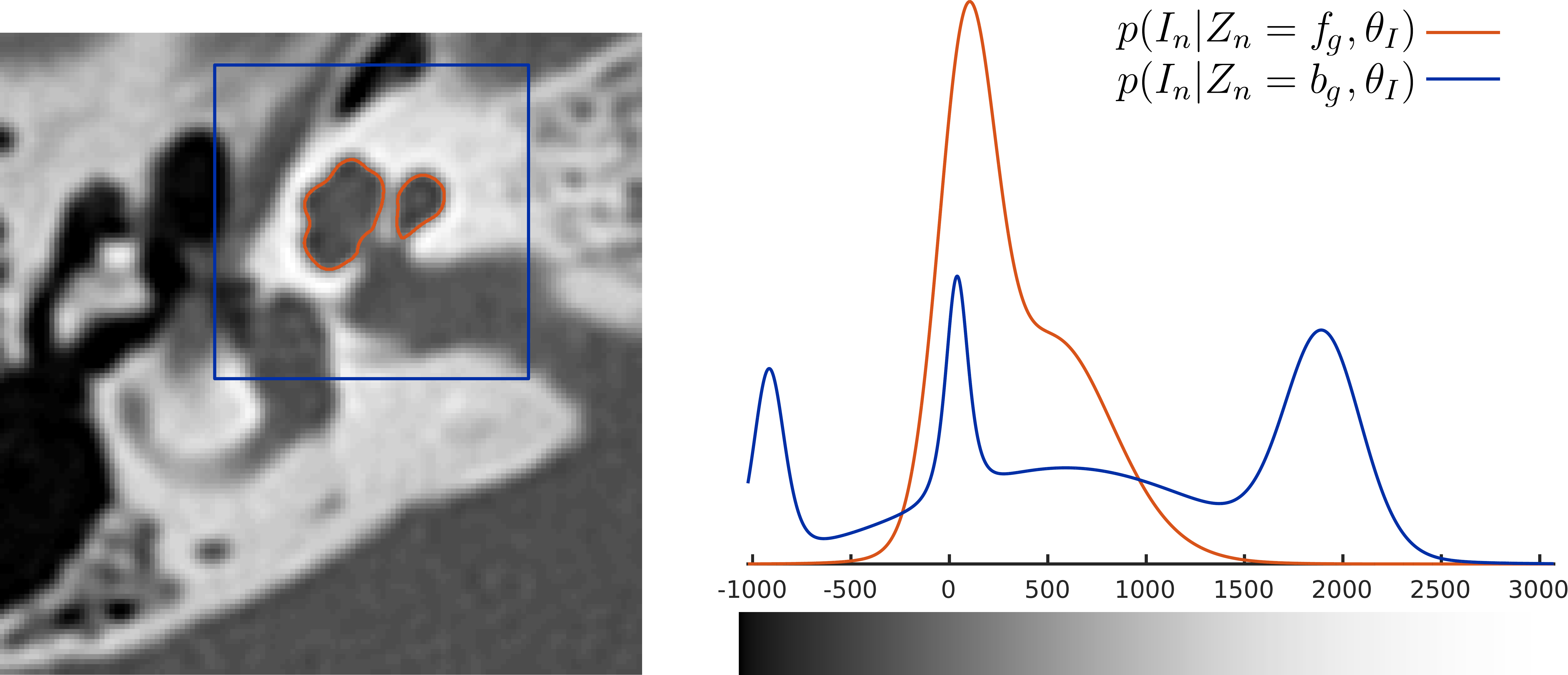}
	\caption[Example of intensity probability distributions]{Example of intensity probability distributions of the foreground ($\fg$, in red) and the background ($\bg$, in blue) as functions of the Hounsfield unit.}
	\label{fig:intensity_model}
\end{figure}
For the cochlea segmentation in CT images, we propose an unsupervised approach based on 
mixture of mixtures of Student's $t$-distributions, i.e. each background and foreground regions are described as mixtures of Student's $t$-distributions. Those $t$-distributions are generalized Gaussian distributions with heavy tails and lead to more robust estimations than GMM since they are less sensitive to extreme intensity values \citep{Peel2000}.
In this context, the probability of observing intensity $I_n$ knowing the label $Z_n$ is parameterized as : 
\begin{equation}
p(I_n|Z_n = k,\theta_I) = \sum_{m = 1}^{M_k} \pi_m^k \: \tpdf(I_n|\mu^{k}_m,\sigma^{k}_m,\nu^{k}_m)\,,
\end{equation}
where $M_k$ corresponds to the number of mixture components for the class  $k$ and mixture coefficients $\pi_m^k$ are positive and sum to one $\sum_{m=1}^{M_k}\pi_m^k=1$. The mean parameter $\mu^{k}_i$, standard deviation coefficient $\sigma^{k}_i$ and degrees of freedom $\nu^{k}_i$ are parameters of the Student's $t$-distribution defined as :
\begin{equation}
\tpdf(I_n|\mu,\sigma,\nu) = \frac{\Gamma\left(\frac{\nu+1}{2}\right)}{\Gamma\left(\frac{\nu}{2}\right)}\frac{1}{\sqrt{\pi\nu} \sigma}\left( 1 + \frac{(I_n-\mu)^2}{\sigma^2 \nu} \right)^{-\left(\frac{\nu + 1}{2}\right)} \,,
\end{equation}
where $\Gamma(\cdot)$ is the gamma function.  To write the likelihood of  this Student's $t$-distribution mixture of mixtures, we introduce a new categorical
variable $\tau_{nkm}$ which is a binary 1-of-Mk encoding such that $\tau_{nkm} = 1$ if voxel n
belongs to the m-th component of region k, and $\sum_{m=1}^{M_k}  \tau_{nkm} = 1$. The
likelihood then writes as:
\[
p(I_n|Z_n,\tau_n)=\prod_{k=0}^1 \prod_{m=1}^{M_k} \left[\left(\tpdf(I_n|\mu_m^k,\sigma_m^k,\nu_m^k)\right)^{\tau_{nkm}}\right]^{Z_{nk}}
\] The inference is performed with closed-form updates of all parameters~\citep{Peel2000,bishopbook} after writing the Student's $t$-distribution as a Gaussian scale mixture. The total number of parameters to estimate is then $|\parami |=4(M_0+M_1)$.
For the cochlea segmentation problem, we assume that the cochlea region mainly consists of two components ($M_1=2$) :  the  fluid (perilymph and endolymph) component centered around  0~HU and the bony walls centered around  500~HU. For the cochlea background,   we consider  $4$ components ($M_0=4$) centered  around 0~HU (fluid), 2000~HU (bony labyrinth), -1000~HU (air due to pneumatization) and 600~HU (temporal bone). The corresponding initial distribution of intensity in the background and foreground regions are shown in Fig.\ref{fig:intensity_model} and the exact initialization values are provided in \ref{sec;initialIntensity}.
\section{Results}
\label{sec;results}
\subsection{Synthetic Images}

\begin{figure}
\centering
\includegraphics[width=\linewidth]{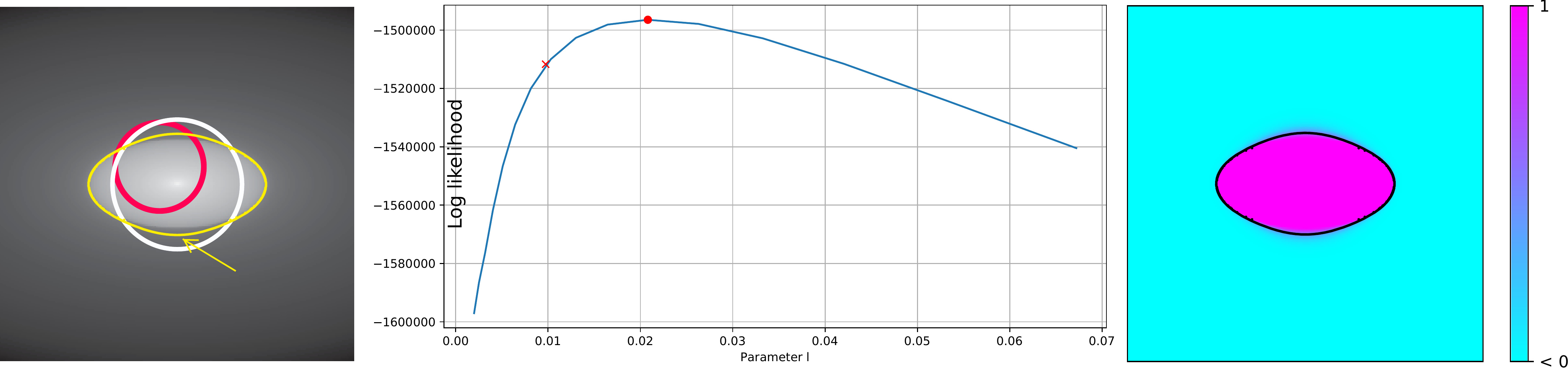}
\caption{(Left) Input Ellipse image fitted with a circle shape : initial  circle (red), final circle (white) and 0.5 isocontour of posterior label probability for optimal value of $\lref$ (yellow);(Middle)  Log likelihood as a function of $\lref$  ; (Right) posterior label probability $p(Z_n=1|\params,\parami)$ for optimal value of $\lref$;}
\label{fig:ellipse}
\end{figure}

We provide a  2D synthetic example to illustrate the influence of the reference length $\lref$ in  the proposed segmentation algorithm.
We consider the segmentation of an ellipse with  Gaussian intensity distribution on both background and foreground  (see Fig.~\ref{fig:ellipse}~(Top Left)) by using a circle prior shape $\shapet(\params,\pos)=\|\pos-\mathbf{C}\|^2-R^2$.  It illustrates the frequent case where the parametric model used as a prior is far simpler than the shape visible in the image. The intensity model consists of two Gaussian distributions initialized with  mean and variance offsets and the circle is parameterized by its center  coordinates and radius. 
The trade-off between imaging information (leading to an ellipse) and prior shape (leading to a circle) is controlled by the $\lref$ parameter. The log-likelihood as a function of $\lref$ exhibits a single maximum for $\lref=\lopt=0.021$ (Fig.~\ref{fig:ellipse}~(Middle)) corresponding to the white circle in Fig.~\ref{fig:ellipse}~(Left) and to the posterior label distribution in Fig.~\ref{fig:ellipse}~(Right). The resulting segmentation is the isocontour $p(Z_n=1)=0.5$, displayed as a yellow curve in Fig.~\ref{fig:ellipse}~(Left), which closely matches the elliptic shape except at its flat part (see arrow). This optimal value of  $\lref$ corresponds to a configuration where the area of the circle is roughly equal to the area of the ellipse.
A value of $\lref<\lopt$ leads to  isocontours $p(Z_n=e_1)=0.5$ that fit more closely the ellipse whereas  $\lref>\lopt$ leads to isocontours that look more like a circle.

\subsection{Inner Ear Datasets} 
\begin{figure}[h!]
    \centering
    \includegraphics[width=0.45\textwidth]{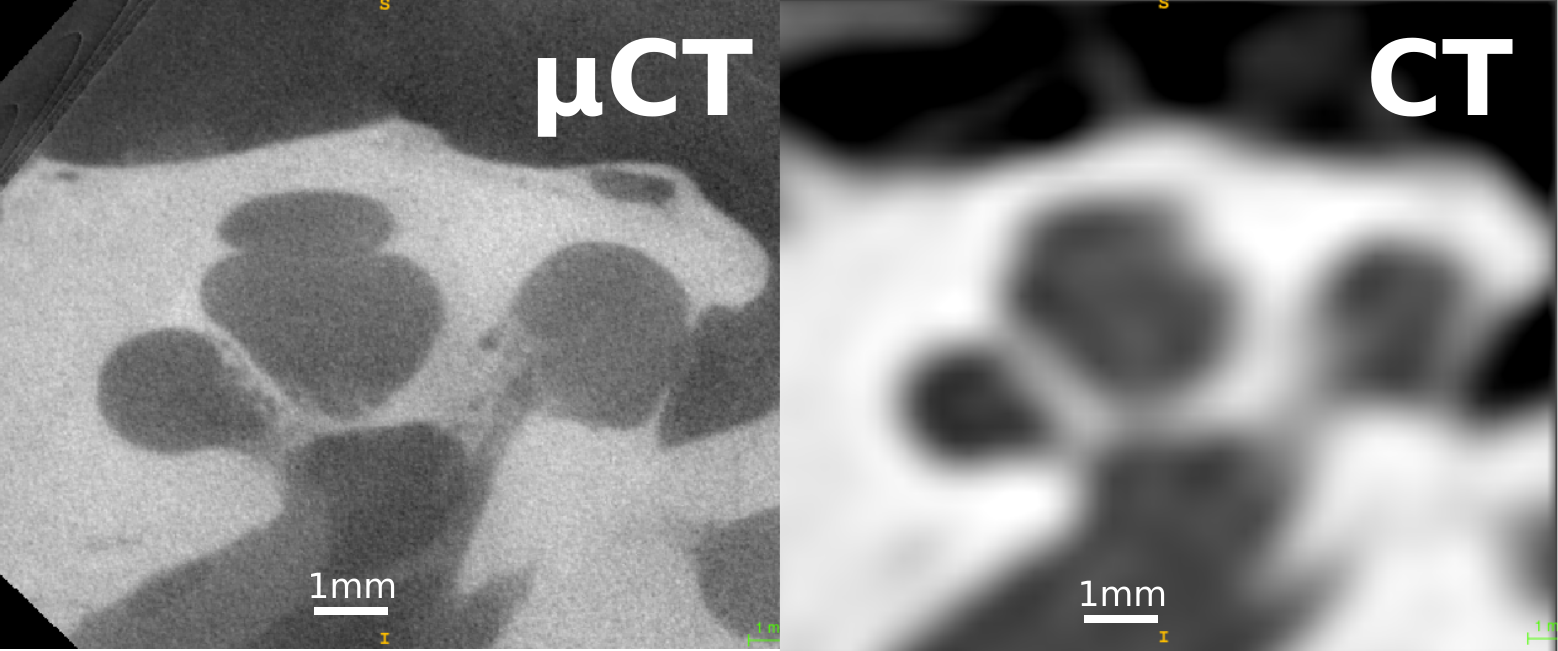}
    \caption{A visual comparison of imaging resolution between the $\mu CT$ and conventional $CT$ for cochlea imaging.}
    \label{fig:muCT}
\end{figure}

The evaluation of the proposed approach is studied on 3 different datasets.
\paragraph{Dataset $\#1$} includes spiral CT temporal bone images of 210 patients from the radiology department of Nice University  Hospital of size $512\times 512 \times 178$ corresponding to a voxel size of  $0.185mm, 0.185mm, 0.25 mm$.  They  have then been  registered to  a reference image via an automatic pyramidal blocking-matching (APBM) algorithm~\citep{Ourselin2000} from the software MedInria \citep{Toussaint2007} followed by an image reformatting around the cochlea to the dimension $(60, 50, 50)$ with isotropic voxel size of $0.2mm$.
The relatively robust registration provides a rough alignment of the cochlea visible in the input image with a cochlea reference frame. From that dataset,  5 CT images were manually segmented by an ENT surgeon (see section~\ref{subsec;semiquant}).
\paragraph{Dataset $\#2$} includes 9 cadaveric cochlea spiral CT images acquired at the face and neck institute  at Nice University  Hospital with the same size and voxel spacing as dataset $\#1$. In addition to CT images,  high resolution  X-ray microtomography (\aka $\mu CT$) images with dimension of $(1035, 800, 1095)$ and isotropic voxel spacing of $0.02479mm$ were acquired each subject. The 9 $\mu CT$ and spiral CT images have been registered together as shown in Fig~\ref{fig:muCT}   and reformatted around the cochlea to the same physical size as for dataset $\#1$ (i.e. $12mm, 10mm, 10mm$). The cochlea and its two scala have been segmented on both CT and $\mu CT$ images by an ENT surgeon with a semi-interactive tool~\citep{Criminisi2008}. The high resolution $\mu CT$ masks serve as ground truth information for the location of the cochlea.
\paragraph{Dataset $\#3$} is a human bony labyrinth dataset \citep{wimmerData} which includes 22 bony labyrinth CT images and their corresponding $\mu$CT images having  isometric voxel size respectively of  $0.1562 mm$  and  $0.0607 mm$. Those images were preprocessed and reformatted as for dataset \#2 and also contains manually segmented cochlea masks. 

\subsection{Quantitative evaluation of segmentation on post-mortem \texorpdfstring{$\mu CT$}{mCT}/CT datasets \#2 and  \#3}

{\bfseries Baseline Approach:} 
We have implemented a 3D atlas based segmentation approach and applied it on  dataset \#2 and \#3 to get a baseline accuracy in terms of Dice score. To this end, we randomly select one image from each dataset as template image and for each input image we perform a multiscale demons deformable registration  \citep{vercauteren:inria-00616035}  (as implemented in SimpleITK 1.0.1) to estimate the deformation field. The segmented mask of the template is deformed to match the target image. 
 The average Dice scores are 0.63 for dataset~\#2 and 0.68 for dataset~\#3.

{\bfseries Logistic Shape Model Inference:} In all cases, the deformable shape parameters $\paramsd$ were initialized as $(a = 4, b = 0.15, \alpha = 0.6, \varphi = 0.2)$ and the pose parameters were set to zero. The $\lref$ value was set between 0.1 and 0.3 (see section~\ref{subsec;reflength})  and
the stopping  condition  is $\frac{\Delta \theta}{\theta} < 0.1$, thus stopping when   parameter updates are less than 10\% of the parameter values.

{\bfseries Computational efficiency:} 
We analyze the computational cost of several alternative formulations of our algorithm.  More precisely, in Table~\ref{tab:evaluation} we compare the computational time  of three different implementations of our approach that differ by the choice of the quasi-Newton optimization method in the MS-step (BFGS vs Gauss-Newton) and by the algorithm used for generating signed distance maps ( VTK based vs deep learning based).  The various  algorithms was applied on the 9 images of dataset \#2 and ran on a Dell Precision 7520 computer. It is clear that the Gauss-Newton method described in Algorithm~\ref{alg:msstep} is far more efficient since it uses a much better approximation of the Hessian matrix than in the generic quasi Newton approach. Furthermore, as expected, the trained deep learning method leads to a speedup factor greater than 3.
\begin{table}
\centering
\caption{Computational efficiency  proposed methods}
\label{tab:evaluation}
\begin{tabular}{cccc} 
\hline
                   & BFGS & VTK SDM   & \textcolor[rgb]{0.2,0.2,0.2}{DLSDM}   \\ 
\hline
Mean Comput. Time          & 12h15min  & 43min & 16min                                          \\
\hline
\end{tabular}
\label{tab:avg_dice_ct}
\end{table}

\begin{table*}
\centering
\caption{Performance metrics obtained on dataset \#2 and \#3.}
\label{tab:seg_results}
\begin{tabular}{cccccccc} 
\hline
\multicolumn{2}{c}{\multirow{3}{*}{\textbf{Compared Labels}}}                                                 & \multicolumn{2}{c}{\multirow{2}{*}{\textbf{Dice Score}}} & \multicolumn{4}{c}{\begin{tabular}[c]{@{}c@{}}\textbf{Symmetric Hausdorff Distance }\\\textbf{(voxel size 0.2 mm)}\end{tabular}}  \\ 
\cline{5-8}
\multicolumn{2}{c}{}                                                                                          & \multicolumn{2}{c}{}                                     & \multicolumn{2}{c}{\textbf{\textbf{Dataset~\#2}}} & \multicolumn{2}{c}{\textbf{\textbf{\textbf{\textbf{ Dataset~\#3}}}}}                   \\ 
\cline{3-8}
\multicolumn{2}{c}{}                                                                                          & \textbf{Dataset \#2 } & \textbf{Dataset \#3 }            & \textbf{~95\%} & \textbf{100\%}                    & \textbf{\textbf{~95\%}} & \textbf{\textbf{100\%}}                                      \\ 
\hline
\multirow{2}{*}{CT Manual 
}                                                          & SSI 
& 0.74 $\pm$ 0.02       & 0.77 $\pm$0.023                  & 0.53           & 1.04                              & 0.70                    & 1.91                                                         \\
                                                                                         & SROI 
                                                                                         & 0.85 $\pm$ 0.011      & 0.91 $\pm$0.015                  & 0.34           & 0.82                              & 0.36                    & 1.68                                                         \\ 
\hline
\multirow{3}{*}{\begin{tabular}[c]{@{}c@{}}$\mu$CT Manual\\ \end{tabular}} & SSI 
& 0.67 $\pm$ 0.024      & 0.76 $\pm$0.068                  & 0.68           & 1.48                              & 0.67                    & 1.96                                                         \\
                                                                                         & SROI 
                                                                                         & 0.81 $\pm$ 0.04       & 0.91$\pm$ 0.019                  & 0.50           & 1.31                              & 0.36                    & 1.68                                                         \\
                                                                                         & CT Manual
                                                                                         & 0.70 $\pm$ 0.084      & 0.93$\pm$ 0.021                  & 0.50           & 1.34                              & 0.19                    & 0.74                                                         \\
\hline
\end{tabular}
\end{table*}

{\bfseries Influence of the reference length} 
\label{subsec;reflength}
To assess the influence of the hyper parameter reference length: $l_{ref}$, we analyse the variation of the final Dice score for various reference lengths based on one image of dataset \#2. The results are shown in Table~\ref{tab:lref}. We see that the reference length within the range $[0.05mm,0.25mm]$ has a relatively small influence on the Dice score. 
To minimize the time of computation, we do not optimize the reference length through a greedy search  but simply set its value to 0.1 for dataset \#1 and \#2 and 0.3 for dataset \#3 for shape fitting. To compute the final hard segmentation we use a fixed reference length of 0.25.
\begin{table}
\centering
\caption{Influence of the hyper parameter: $l_{ref}$ for the segmentation accuracy.}
\label{tab:lref}
\begin{tabular}{cccccc} 
\hline
\textbf{ Ref. Length } & 0.05 & 0.1  & 0.15 & 0.2  & 0.25  \\
\textbf{Dice Score }   & 0.84 & 0.85 & 0.85 & 0.82 & 0.84  \\
\hline
\end{tabular}
\end{table}

{\bfseries Robustness analysis }
To study the robustness of the method, we randomly initialize the cochlea shape parameters by performing a random uniform sampling within their defined value range. Based on 10 initial random samples, we computed the average Dice score for one image of dataset \#2 and obtained a mean Dice score of $0.81$ $\pm 0.1$ ( respectively of $0.68$ $\pm 0.23$  ) for the  Gauss-Newton  method (resp. the BFGS method). This clearly shows the increased robustness with respect to initial shape values  obtained by the Gauss-Newton optimization of the MS-step.

{\bfseries Evaluation on CT and $\mu$CT images:} 
Datasets $\#2$ and $\#3$ include both CT and $\mu$CT images of the same subject that have been registered to each other. Furthermore the cochlea was manually or semi-automatically segmented by an expert on both modalities such that we can use those two binary maps to evaluate the accuracy of the algorithm applied on the CT image. The cochlea binary map from  high resolution $\mu$CT images have been downsampled and represent a more reliable ground truth than the manual segmentation performed on the CT images. 

The proposed algorithm using Gauss-Newton optimization and deep-learning generation of signed distance maps was applied on the $9+22$ CT images of the two datasets.  Fig.~\ref{fig:segmentationRes}~(Right)  shows the segmented cochlea in red, the associated shape model, and its evolution during the MS step. Clearly, we see that the resulting segmentation is strongly constrained by the shape model.

In Table~\ref{tab:seg_results}, we provide two metrics between pairs of binary masks : the Dice score and the 95\% and 100\% symmetric Hausdorff distance (HD) (computed as the average of two  distances). Furthermore, we compare the segmentations produced by the posterior label probability (SROI for $p(Z_n|I_n,\params,\parami)=0.5$) and the ones produced by the shape model only (SSI for $p(Z_n|\params)=0.5$) with both manual segmentations obtained on CT and  $\mu$CT images. To measure the uncertainty in the manual CT segmentation, we also evaluate the metrics between both CT and  $\mu$CT manual mask images.

The logistic shape model framework produces good segmentation results on both datasets (Dice scores of $0.81$ and $0.91$) and even slightly outperforms the manual CT segmentation on dataset$\#2$ ($0.81$ vs $0.7$) which is far more challenging dataset~$\#3$. The segmented shape instances produced by the shape model are not as accurate as the SROI for the cochlea segmentation (lower Dice score and larger HD). This confirms that the parametric geometric cochlea model  is a simplified representation of the cochlea anatomy. Finally, the metrics between the 2 manual segmentations on dataset~$\#2$ (DSC of  $0.7$ with a 95\% HD of 0.5mm) shows the difficulty of performing a manual segmentation of the cochlea due to its limited size and contrast.  
\subsection{Semi-quantitative analysis  of segmentation on clinical dataset \#1} \label{subsec;semiquant}

We ran the segmentation framework (with DLSDM) on the 210 CT of  dataset~$\#1$ on a Dell 6145 and 6420 CPU clusters.

\begin{figure}[h!]
    \centering
    \includegraphics[width=0.45\textwidth]{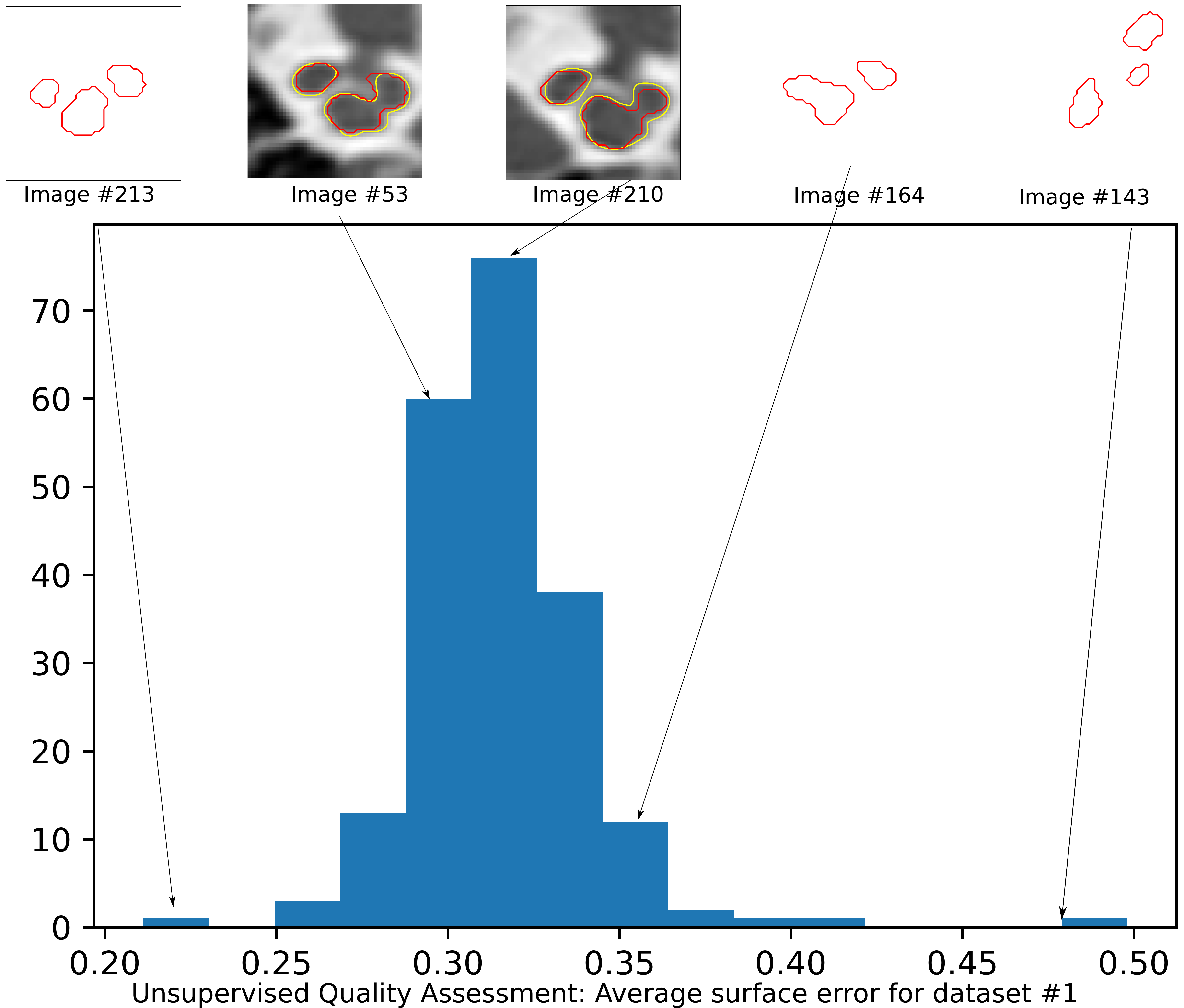}
    \caption{Average surface error of segmentations generated from dataset \#1 resulting from the unsupervised quality control.  Red contours correspond to the manual ground truth while yellow ones are segmentation outputs.}
    \label{fig:unsegqc}
\end{figure}

{\bfseries Unsupervised quality control and semi-quantitative evaluation}
As  manual segmentations of the 210 images are not available, we propose instead an original approach to estimate our algorithm's performance while minimizing the manual annotation effort. First, we  apply  the unsupervised quality control  algorithm of Audelan \emph{et al.}~\citep{audelan} on the whole dataset in order to sort the 210 segmentations according to their hypothesized performance. More precisely, this quality control algorithm computes for each image segmentation, an average distance between the segmentation provided by our algorithm and a segmentation produced by a simple generic probabilistic method. We can then generate  an histogram of such average surface error (ASE) in Fig.~\ref{fig:unsegqc}. Segmentations having a low ASE correspond to those having good intensity contrast across their boundaries while those on the right tail of the distributions are considered as more challenging and suspect of including segmentation errors.   

The histogram exhibits a bell shape with few outliers on its right and left tails.  Furthermore, we have manually checked that this unsupervised quality control algorithms worked well on this dataset with visually better segmentations localized on the left side of the histogram. To estimate the relation between the ASE and the Dice score, we picked 5  images  in order to sample the histogram at different levels of ASE corresponding to images  \#213, \#210, \#53, \#264 and \#143 (see Fig.\ref{fig:unsegqc}) in ascending order of ASE. Those 5 CT images were manually segmented by an ENT surgeon and the Dice scores of the segmentation produced by our algorithm was reported in  Table~\ref{tab:segmentatioUnSegQC}. We see that the Dice score decreases as the ASE increases which indicates that the ASE may be a proper surrogate for the segmentation performance. The cochlea in image \#143 was indeed found to be  an outlier in terms of shape probably due to a patient malformation. Inspired by \citep{audelan}, we can make the hypothesis that ASE a good proxy for the Dice score as there is a monotonic relation between ASE and Dice. On this basis, we can extrapolate that the median Dice score over the whole dataset is probably above 0.82. Yet,  a more thorough study with far more manual segmentations is necessary to be less speculative about the actual performance on clinical CT data. 




\begin{table}
\centering
\caption{Dice score for selected segmentation samples from dataset \#1 based on the histogram of Fig.\ref{fig:unsegqc}. The ASE are got from automatic quality control algorithm and the DICE score are computed based on manual segmentation.}
\label{tab:segmentatioUnSegQC}
\begin{tabular}{lllllll}
\textbf{Patient ID } & 213  & 210  & 53   & 164  & 143   \\ 
\hline
\textbf{DICE Score } & 0.84 & 0.84 & 0.84 & 0.76 & 0.45  \\
\textbf{ASE }        & 0.21 & 0.30 & 0.32 & 0.36 & 0.50  \\
\hline 
\end{tabular}
\end{table}

\begin{figure*}[h!]
    \centering
    \includegraphics[width=\textwidth]{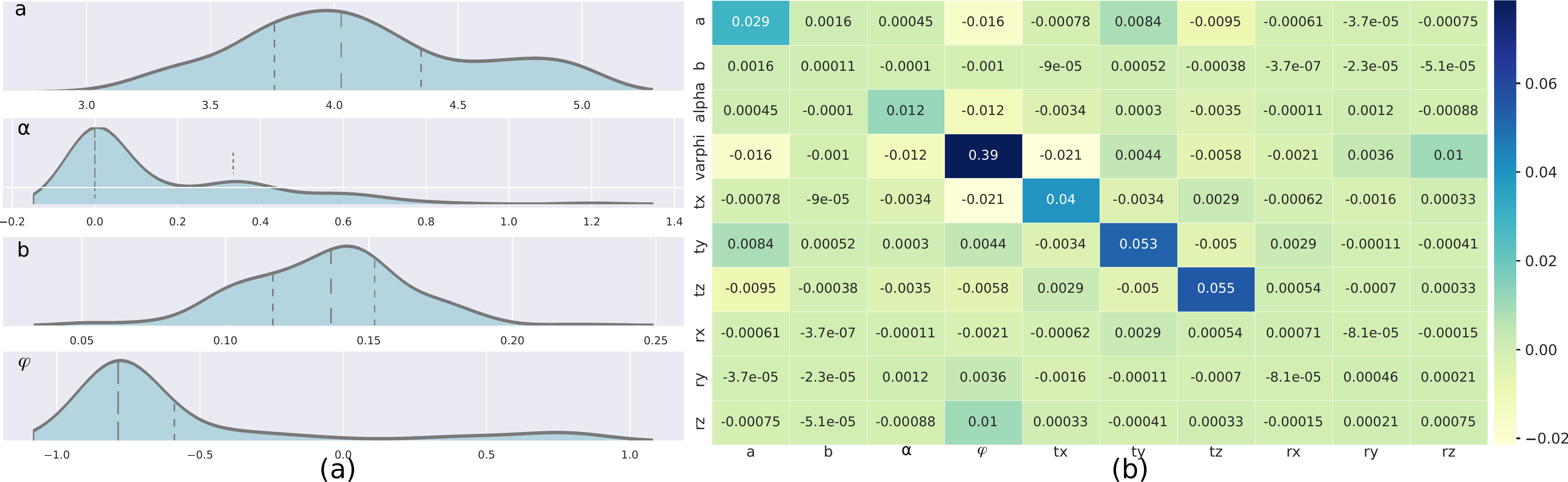}
    \caption{Distribution plots for shape parameters variance.  (b) Average covariance matrix of the 10 shape parameters.}
    \label{fig:fittingRes}
\end{figure*}

{\bfseries Parameter analysis}
The application of the algorithm on dataset \#1 resulted in the estimation of 10*210 shape parameters with 210 covariance matrices $\cov_{\params}^{\star}$. 
 In  Fig.~\ref{fig:fittingRes}(a) the histograms of the 4 deformable shape parameters are displayed in green. Interestingly, the $a$ and $\alpha$ parameters exhibit a bimodal distribution for which a simple explanation may be provided. Indeed, the left highest mode is probably corresponding to  straight centerline profiles whereas the rightmost mode may be associated with the "rollercoaster" longitudinal profiles~\citep{Avci2014}.
 In Fig.~\ref{fig:fittingRes}(b) the average $10\times10$ covariance computed as the log-Euclidean mean~\citep{Arsigny:Siam:07}  is displayed  showing potential correlations between pose and deformable parameters. Shape parameter $b$ corresponding the exponent of the logarithmic center line curve has a particularly low variance such that it can be well estimated from the data. Conversely the phase parameter $\varphi$ has much greater variance and is harder to estimate in average. The extraction of the eigenvectors of that covariance matrix confirms that most parameters are independent 
 with other except for parameter $a$ which is a bit correlated with the $\varphi$ parameter and the translation term $ty$. The relative independence of the parameters for shape fitting indicates that we do not have an overparametrization of the cochla shape. 
 
\begin{figure}
\centering
\includegraphics[width=0.47\textwidth]{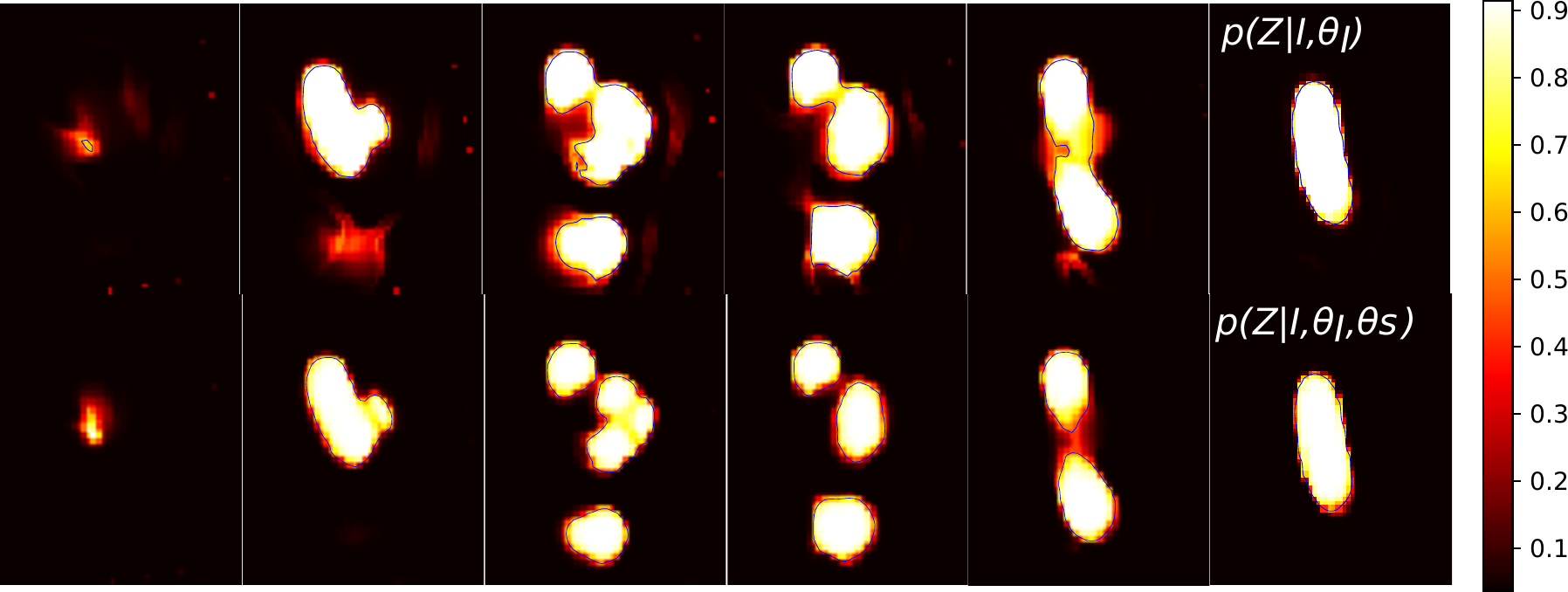}
\caption{Marginal posterior probability $p(Z|I,\parami)$ (Top) versus posterior probability $p(Z|I,\parami,\params)$ (Bottom) computed on patient \#1 of dataset \#1.}
\label{fig:uncer}
\end{figure}

{\bfseries Uncertainty Segmentation analysis} 
 The estimated covariance matrix  $\cov_{\params}^{\star}$ can be used for studying the uncertainty of the output segmentation. We sampled 100 times the multivariate Gaussian approximate posterior distribution of the parameters $p(\params|I)\approx \mathcal{N}(\params^\star;\cov^\star)$ and generated accordingly  100 random posterior labels $p(Z|I,\params,\parami)$ that are then averaged to  estimate with Monte-Carlo sampling the marginal posterior $p(Z|I,\parami)=\int_{\euc^{|\params|}} p(Z|I,\params,\parami) p(\params|I)~d\params$. In   Fig~\ref{fig:uncer} we show several slices of the resulting probability maps with the 0.5 level curve together with the posterior probability $p(Z|I,\params^\star,\parami)$ obtained with the most likely shape parameter $\params^\star$. We see that we have a much larger uncertainty in the resulting segmentation when accounting for the uncertainty in the shape parameters than without them. This is a far better approximation of the true uncertainty $p(Z|I)$ than the posterior label probability $p(Z|I,\params,\parami)$.

\subsection{Comparison with the state-of-the-art}
We consider below the prior work on cochlea segmentation evaluated on clinical CT images while discarding the literature on the segmentation of $\mu$CT images~\citep{Kjer2014a,RuizPujadas2016,RuizPujadas2016a} or of the scala tympani and vestibuli located inside the cochlea~\citep{Noble2012,Noble2013}. Table~\ref{tab:previous} summarises the relevant publications on  cochlea segmentation that are split into unsupervised and supervised methods. The former approaches are mostly based on  cochlear shape fitting based on template image registration~\citep{Baker2005},  parametric shape model~\citep{Baker2008}. The supervised methods are based on
statistical deformation models~\citep{PujadasKjer} and deep learning~\citep{DICE90CNN,DICEUNET90,DICE90}. 

Quantitative comparison of performances is not straightforward due to  differences in image modality (CT, $\mu CT$ or ultra high resolution CT), in metrics (Dice, precision, mean surface error), in subject population (cadaveric vs patient) but also in the target anatomical structures (cochlea vs cochlea labyrinth). In most cases, cochlea segmentation from $\mu CT$ images are used as ground truth information and a direct comparison between our work with~\citep{DICEUNET90} is possible since they  used a subset of dataset \#3 which is a public database~\citep{wimmerData}. We see that our unsupervised approach performs as well as the supervised methods with 
Dice scores in the range $[0.85,0.91]$ and outperforms previous unsupervised methods.

\begin{landscape}
\begin{table*}
\centering
\caption{Performances of prior work on cochlea segmentation. \em{NL} (resp. \em{NT}) indicates the number of training (resp. testing) images. \em{Unsup} (resp. \em{Sup}) refers to unsupervised (resp. supervised) learning  methods. }
\label{tab:previous}
\begin{tabular}{ccccccc}
\textbf{Method Group} & \begin{tabular}[c]{@{}c@{}}\textbf{Study}\\\end{tabular}                              & \textbf{Comparison } & \textbf{Metrics }                                                         &                                                                                             & \begin{tabular}[c]{@{}c@{}}\textbf{Proposed}\\\textbf{ method}\\\textbf{(Dataset\#2 N=9) }\end{tabular} & \begin{tabular}[c]{@{}c@{}}\textbf{Proposed}\\\textbf{ method}\\\textbf{(Dataset\#3 N=22) }\end{tabular}  \\ 
\hline
UnSup                 & \begin{tabular}[c]{@{}c@{}}\citet{Baker2008}\\ (NT= 4)\end{tabular}                           & CT                   & Precision                                                                 & \begin{tabular}[c]{@{}c@{}}0.72\\ $\pm$ 0.09\end{tabular}                                   & \begin{tabular}[c]{@{}c@{}}\textbf{0.75}\\ \textbf{$\pm$ 0.03}\end{tabular}                             & \begin{tabular}[c]{@{}c@{}}\textbf{0.83}\\\textbf{$\pm$0.03 }\end{tabular}                                \\ 
\cline{2-7}
UnSup                 & \begin{tabular}[c]{@{}c@{}}\citet{Kjer2015}\\ (NT = 2)\end{tabular}                          & post-mortem $\mu$CT & \begin{tabular}[c]{@{}c@{}}Mean\\($\pm$1 std)\\surface error\end{tabular} & \begin{tabular}[c]{@{}c@{}}0.22\\$\pm$0.17\end{tabular}                                     & \begin{tabular}[c]{@{}c@{}}\textbf{0.11}\\\textbf{$\pm$0.06}\end{tabular}                               & \begin{tabular}[c]{@{}c@{}}\textbf{0.063}\\\textbf{$\pm$0.03}\end{tabular}                                \\ 
\hline
Sup                   & \begin{tabular}[c]{@{}c@{}}\citet{Kjer2017PatientspecificEO}\\(NL = 18 /NT = 14)\end{tabular} & post-mortem CT                   & Dice                                                                      & 0.88                                                                                        &                                                                                                         &                                                                                                           \\
\cline{2-5}
Sup                   & \begin{tabular}[c]{@{}c@{}}\citet{DICE90}\\(NL=48/NT = 75)\end{tabular}                       & \begin{tabular}[c]{@{}c@{}}  Ultra-high \\  Resolution CT\end{tabular}                 & Dice                                                                      & \begin{tabular}[c]{@{}c@{}}0.90\\ \textbf{$\pm$} 0.03\end{tabular}                          &                                                                                                         &                                                                                                           \\ 
\cline{2-5}
Sup                   & \begin{tabular}[c]{@{}c@{}}\citet{DICE90CNN}\\(NL=24/NT=6)\end{tabular}                      & Cochlea Labyrinth      & Dice                                                                      & 0.90                                                                                        & \multirow{4}{*}{\begin{tabular}[c]{@{}c@{}}0.85\\$\pm$0.011\end{tabular}}                               & \multirow{4}{*}{\begin{tabular}[c]{@{}c@{}}\textbf{0.91}\\\textbf{$\pm$0.03}\end{tabular}}                \\ 
\cline{2-5}
Sup                   & \begin{tabular}[c]{@{}c@{}}\citet{DICEUNET90}\\(NL + NT = 17)\end{tabular}                      & post-mortem CT                   & Dice                                                                      & \begin{tabular}[c]{@{}c@{}}\textbf{0.90}\\\textbf{$\pm$\textbf{\textbf{0.07}}}\end{tabular} &                                                                                                         &                                                                                                           \\ 
\hline
\end{tabular}
\end{table*}
\end{landscape}

\section{Discussion}

The proposed approach relies on the definition of a generic shape function $\shapet(\params,\pos)$ which can be for instance a statistical shape model, a deformation image template, or an implicit shape equation. In the case of the cochlea, it was defined as a signed distance function of a parametric shape model $\sdm(\shape(\params),\pos)$. This specific choice makes the computation of the shape function and its gradient fairly costly, despite the use of a fully supervised dedicated neural network (DLSDM). There are several ways to optimize its  computation time. One could for instance  use a supervised appearance model such as a trained neural network which would remove all  MI steps in the EM algorithm and would decrease by at least a factor 2 the time of computation. Another way is to use an implicit shape model  $\shape(\params,\pos)=0$ for instance based on  statistical level sets~\citep{Tsai2003}. The cochlea segmentation example provided in this paper relies on a fully interpretable intensity and shape parameters at the expense of its computational efficiency. Yet, one could train a deep neural regressor for predicting cochlea shape parameters and segmentation by using  the segmentations generated by the proposed framework as training set.

For the cochlea segmentation, excellent results were obtained on cadaveric CT images similarly to the supervised methods. Furthermore, to the best of our knowledge, we introduced a first semi-quantitative assessment of cochlea segmentations on clinical CT images acquired on more than 200 patients. However, for a complete study, one would need to assess thoroughly the inter-rater variability of those manual segmentations and ideally combine them with other high resolution image modalities. Finally, an interesting extension of this work would be to segment the scala vestibuli and tympani  in addition to the cochlea.
\section{Conclusion}
In this paper, we have presented a new probabilistic generative approach for combining shape and intensity models for image segmentation. The resulting segmentation is an interpretable compromise between a fidelity to a parametric shape space (captured in the prior distribution) and  an appearance model (captured in the likelihood distribution). The proposed method goes well beyond the concept of shape fitting since it also provides an approximation to the posterior distribution of shape parameters. The use of a logistic shape model allows to control the trade-off between appearance and shape with a single parameter: the reference length. When applied to the recovery of cochlea structures from CT images, we were able to provide accurate segmentations with meaningful shape parameter distributions. Furthermore, we have shown how the approximate shape parameter posterior distribution can be exploited to provide realistic uncertainty maps.

An interesting application of the proposed approach is to perform model selection with Bayes factors, in order to estimate the optimal complexity of a parametric shape model for a given image segmentation task.
Future work will also explore the application of this framework to other shape representations than explicit parametric shape models in order to find a reasonable trade-off between computational efficiency and interpretability of shape parameters. For instance, in statistical deformation models~\citep{sdms}, the computation of  shape function gradient $\nabla_{\params}\shapet(\params,\pos)$ is straightforward, but its shape parameters may not be meaningful besides the first modes. 

\appendix
\section{Gradient of shape function} 
\label{appendixgradientshape}
In this section, we detail the computation of the shape function gradient $\nabla_{\params} \shapet(\params,\pos)$ when rigid and deformable shape parameters are considered.  More precisely, writing the  parameters controlling the non-rigid deformation as $\paramsd$, the shape function writes as  $\shapet(\paramsd,\rot\pos_n+\trans)$. The rotation matrix $\rot$ is parameterized with rotation vector $\mathbf{r}$, whose norm is the rotation angle and whose direction is the rotation axis. The gradients with respect to the translation and rotation vectors are then given in closed form as :
\begin{align*}
\nabla_{\trans}\shapet(\paramsd,\rot\pos_n+\trans)
&=\nabla_{x}\shapet(\paramsd,\rot\pos_n+\trans)\\
\nabla_{\rotv}\shapet(\paramsd,\rot\pos_n+\trans)
&=\left (-\rot S_{\pos_n}\frac{\rotv\rotv^T+((\rot)^T-\identity_3)S_{\rotv})}{\|\rotv\|^2}\right)^T \\
&\nabla_{x}\shapet(\paramsd,\rot\pos_n+\trans)
\end{align*}
where $\nabla_{x}$ is the spatial gradient, $S_\pos$ is the $3x3$ anti-symmetric matrix associated with vector $\pos$. For a deformable parameter $\paramsd$, if the shape function is not given in an analytical form as it is the case for parametric shapes,  the shape function gradient can be computed with finite differences based on a parameter increment $\delta\paramsd^i$ :
\begin{align*}
\nabla_{\paramsd^i}\shapet(\paramsd,\rot\pos_n+\trans)&= \frac{1}{2\delta\paramsd^i}\left(\shapet(\paramsd+\delta\paramsd^i),\rot\pos_n+\trans)\right.\\&\left.-\shapet(\paramsd-\delta\paramsd^i,\rot\pos_n+\trans)\right)
\end{align*}

\section{Cochlea Shape Model}
\label{appendixShape}

We are interested in the cochlea structure in CT images
which is  defined as a generalized cylinder, i.e. as cross-sections swept along a centerline. 

{\bfseries Centerline}
The centerline is parameterized in a
cylindrical coordinate system  by its \textit{radial} $r(\theta_c)$ and \textit{longitudinal} $z(\theta_c)$ components. The  range of polar angle $\theta_c$ is $[0,\theta_{\max}]$ where $\theta_{\max}$ is the maximum polar angle controlling the total number of cochlear turns. 


The radial component is defined piecewise with a polynomial function and a logarithmic function of the angular coordinate $\theta_c$ in the cylindrical coordinate system as:
\begin{equation}
r(\theta_c) =
\begin{cases}
p_2 \theta_c^2 + p_1 \theta_c + p_0 & \mbox{if } \theta_c < \theta_0 \\ a e^{-b \theta_c} & \mbox{if } \theta_c \geq \theta_0
\end{cases}
\end{equation}


where $\theta_0=5\pi/6$ and $p_0=$ 5 mm. Furthermore to obtain a continously differentiable curve, we set : 
\begin{equation}
\begin{aligned}
p_2 &= \frac{C_1\theta_0 - C_2 + p_0}{\theta_0^2} & & p_1 = \frac{-C_1\theta_0 + 2C_2 + 2p_0}{\theta_0}\\
C_2 &= a e^{-b \theta_0} & & C_1 = -C_2 b\,.
\end{aligned}
\end{equation}


The longitudinal component of the centerline is the sum of an exponentially damped sinusoidal and a linear function: 
\begin{equation}
z(\theta_c) =
\begin{cases}
\alpha e^{-\beta \theta_c} \cos(\theta_c + \phi) + q_1 \theta_c &  \text{if } \theta_c < \theta_1 \\ a_2 \theta_c^2 + a_1 \theta_c + a_0 & \text{if } \theta_c \geq \theta_1
\end{cases}\,,
\label{eq:z_new}
\end{equation}
where $\beta$ = 0.2 rad$^{-1}$, $q_1$ = 0.225 mm.rad$^{-1}$ and $\theta_1 = \theta_{\max} - \pi$. The polynomial function is used to flatten out the last half turn so that $\mathrm{d}z(\theta)$/$\mathrm{d}\theta |_{\theta = \theta_{\max}} = 0$ and similarly $a_2$, $a_1$, $a_0$ are set to obtain a continuously differentiable curve.

{\bfseries Cross-Sections}
The cross-sections are modeled by a closed \textit{planar shape} on which a varying \textit{affine transformation} is applied along the centerline.
The scala tympani and the scala vestibuli are modeled with two half pseudo-cardioids while the cochlear cross-section corresponds to the minimal circumscribed ellipse of the union of the tympanic and vestibular cross-sections. The affine transform of cross-sections is parameterized by a \textit{rotation}, and a \textit{width} and  \textit{height scalings}.
All cross-sectional parameters are fixed because their variability was found to be small compared to the variability  of the centerline. 

{\bfseries Shape parameter vector}
We have chosen a compact description of the cochlea shape to limit as much as possible the correlation between the shape parameters and therefore make them uniquely identifiable. 
Finally, only 10 free parameters are considered in $\params$ :
\begin{itemize}
	\item 6 translation and rotation parameters : $\trans=(tx,ty,tz)$, $\rotv=(rx,ry,rz)$ 
	\item 2 radial component parameters  of the centerline, $a$ and $b$
	\item 2 longitudinal component  parameters, $\alpha$ and $\phi$
\end{itemize}

Note that there are no free cross-section parameters which implies that  $\params$ can be used to define uniquely 
the cochlea.

The prior probabilities on the 10 shape parameters were modeled as being an uniform distribution (uninformative prior)  such that all regularization terms $\log p(\params|\priors)$ can be ignored.

\section{Initialization of intensity parameters}
\label{sec;initialIntensity}
The $4*6=24$ initial intensity parameters for the mixture of Student's $t$ distributions in  datasets $\#1$ and $\#3$ are presented in  Table~\ref{tab:parameters_init}.
\bibliographystyle{model2-names.bst}\biboptions{authoryear}
\bibliography{refs}


\end{document}